%% file: paper.tex
\documentclass[letterpaper, 10 pt, journal, twoside]{ieeetran}

\IEEEoverridecommandlockouts

\usepackage{amsmath} 
\usepackage{amssymb}  
\usepackage[protrusion=false]{microtype}
\usepackage{tensor}
\usepackage{mathtools}
\usepackage{makecell}
\usepackage{todonotes}
\usepackage{siunitx}
\DeclareSIUnit[product-units = single]{\pixel}{px}
\usepackage{booktabs}
\usepackage{mathrsfs}
\usepackage{changepage}
\usepackage{graphicx}

\title{Balancing the Budget: Feature Selection and Tracking for Multi-Camera
Visual-Inertial Odometry}

\author{Lintong Zhang$^{1}$, David Wisth$^{1}$, Marco Camurri$^{1}$, and
Maurice Fallon$^{1}$%
	\thanks{Manuscript received: September, 9, 2021; Revised November, 30,
	2021; Accepted December, 6, 2021.}
	\thanks{This paper was recommended for publication by Editor Sven Behnke
	upon evaluation of the Associate Editor and Reviewers' comments.
	This research was part funded by the EU H2020 Projects THING and Innovate
	UK-funded ORCA Robotics Hub (EP/R026173/1), a Royal Society University
	Research Fellowship (Fallon) and a Google DeepMind studentship (Wisth). It
	has been conducted as part of the ANYmal research community.
	}
	\thanks{$^{1}$Oxford Robotics Institute, Department of Engineering
		Science, University of Oxford, UK {lintong, davidw, mcamurri,
			mfallon}@robots.ox.ac.uk}%
	\thanks{Digital Object Identifier (DOI): see top of this page.}
}

\usepackage[linesnumbered,ruled,vlined]{algorithm2e}

\usepackage{multirow}

\SetKwInput{KwInput}{Input} 
\SetKwInput{KwOutput}{Output} 

\def\includecomments{1}

\usepackage{xcolor} 

\newcommand{\dwisth}[1]{
\if\includecomments1
\todo[color=blue!40,inline]{\textbf{D. Wisth:}~#1}
\fi
}

\newcommand{\lzhang}[1]{
\if\includecomments1
\todo[color=yellow!40,inline]{\textbf{L. Zhang:}~#1}
\fi
}

\newcommand{\mfallon}[1]{
\if\includecomments1
{\color{red}  \textbf{#1}}
\fi
}

\newcommand{\mcamurri}[1]{
\if\includecomments1
\todo[color=green!40,inline]{\textbf{M. Camurri:}#1}
\fi
}

\newcommand{\X}{\mathcal{X}}
\newcommand{\Z}{\mathcal{Z}}
\newcommand{\etalcite}[2]{#1~et~al.~\cite{#2}}

\DeclareMathOperator*{\argmax}{arg\,max}
\DeclareMathOperator*{\argmin}{arg\,min}

\input{shortcuts} 

\makeatletter
\let\NAT@parse\undefined
\makeatother
\usepackage{hyperref}
\begin{document}

\bstctlcite{library:BSTcontrol}

\maketitle


\begin{abstract}
We present a multi-camera visual-inertial odometry system based on factor
graph optimization which estimates motion by using all cameras simultaneously while
retaining a fixed overall feature budget. We focus on motion tracking in
challenging environments, such as narrow corridors, dark spaces with
aggressive motions, and abrupt lighting changes. These scenarios cause
traditional monocular or stereo odometry to fail.
While tracking motion with extra cameras should theoretically prevent
failures, it leads to additional complexity and computational
burden. To overcome these challenges, we introduce two novel methods to improve
multi-camera feature tracking. First, instead of tracking
features separately in each camera, we track features continuously as they move from one camera to another.
This increases accuracy and achieves a more compact factor graph
representation. Second, we select a fixed budget of tracked features across the
cameras to reduce back-end optimization time. We have found that using a
smaller set of informative features can maintain the same tracking accuracy.
Our proposed method was extensively tested using a hardware-synchronized
device consisting of an IMU and four cameras (a front stereo pair
and two lateral) in scenarios including: an underground mine,
large open spaces, and building interiors with narrow stairs and corridors.
Compared to stereo-only state-of-the-art visual-inertial odometry methods, our
approach reduces the drift rate, relative pose error, by up to
\SI{80}{\percent} in translation and \SI{39}{\percent} in rotation.
\end{abstract}

\begin{IEEEkeywords}
		Visual-Inertial SLAM; Omnidirectional Vision; Localization
\end{IEEEkeywords}

\section{Introduction}
\label{sec:introduction}
\begin{figure}[!t]
 \centering
 \includegraphics[width=0.49\columnwidth]{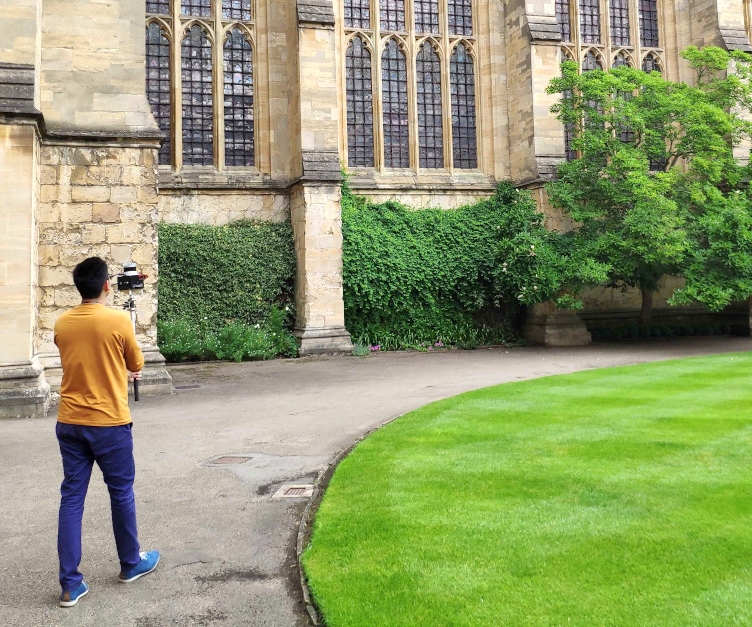}
 \includegraphics[width=0.49\columnwidth]{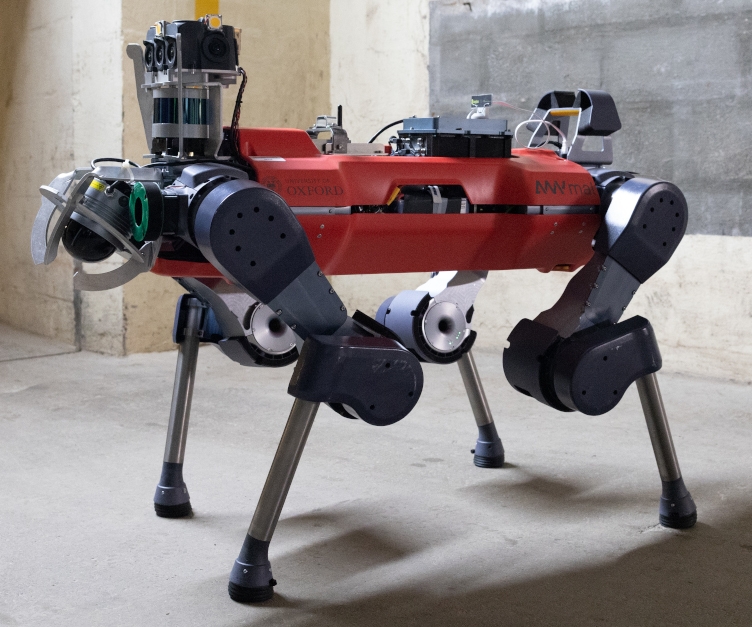}
 \caption{We tested our multi-camera odometry algorithm with data from a
 handheld sensor rig in New College, Oxford (\textit{left}) and the
 ANYmal quadruped \cite{Hutter2016} during preparation for the
 DARPA SubT Challenge \cite{Tranzatto2021} (\textit{right}).\\
 Video: \url{https://youtu.be/cLWeAT72e0U}.}
 \label{fig:anymal-mine}
 \vspace{-3mm}
\end{figure}

\IEEEPARstart{S}{tate} estimation is a fundamental capability required
for autonomous
robot navigation in real-world scenarios. Motion tracking using
cameras is very popular due to their low weight, small form factor, and low
hardware
cost. Specifically, Visual-Inertial Odometry (VIO) methods, which fuse feature
tracking from a camera with estimation from an Inertial Measurement
Unit (IMU), have now become the standard for odometry on Micro Aerial Vehicles
(MAVs) \cite{Forster2017a}. These systems can also be deployed on
a range of platforms, such as handheld rigs, quadrupeds, and vehicles with
applications ranging from indoor and outdoor navigation to underground
exploration \cite{Tranzatto2021} (\Figure \ref{fig:anymal-mine}).

When performing VIO, cameras can track scene information at the frame
rate while the IMU can provide high frequency motion estimates between
keyframes. Although VIO frameworks are generally effective, they are prone to
failure in degenerate scenarios such as over/under exposure, textureless
surfaces, and when there is aggressive motion. Another issue with the minimal
setup composed of just a monocular camera is that this system results in a
single point of failure. Even if IMU measurements can be integrated, the system
can quickly diverge within a few seconds of a feature tracking failure.

Fusing information from additional cameras can greatly
improve robustness and estimation accuracy at the cost of extra complexity and
computational burden. The most obvious benefit of Multi-Camera VIO (MC-VIO) is
redundancy, \eg if one of the cameras suffers from a sudden drop in tracked
features, the other cameras can still estimate motion.
Furthermore, if the cameras are arranged to have overlapping Fields of View
(FoV), a feature could potentially be tracked across the cameras
for as long as it is visible in any camera. Multi-camera solutions have been
presented in the past \cite{Jaekel2020, Seok2020, Eckenhoff2021}, however, no
proposal has studied cross camera feature tracking.

In this paper, we explore how to effectively track features across all the
cameras and how to select the best subset of features to keep the computational
time bounded.

\subsection{Contribution}
\label{sec:contribution}
The main contributions of this work are the following:
\begin{itemize}
\item A novel factor graph formulation that tightly fuses tracked features
from any number of stereo and monocular cameras, along with IMU measurements,
in a single consistent optimization process.
\item A simple and effective method to track features across cameras with
overlapping FoVs to reduce duplicate landmark tracking and improve accuracy.
\item A submatrix feature selection (SFS) scheme that selects the best landmarks
for optimization with a fixed feature budget. This bounds computational time and
achieves the same accuracy compared to using all available features.
\item
Extensive experimental evaluation across a range of scenarios demonstrating
superior robustness, particularly when VIO with an individual camera fails.
\end{itemize}

The proposed algorithm, VILENS Multi-Camera (VILENS-MC), builds upon our
previous VILENS estimation
system \cite{Wisth2021, Wisth2021tro}, by fusing multiple cameras and improving
front-end feature processing.

\section{Related Work}
\label{sec:related-work}
There has been extensive research into monocular and stereo camera VIO.
These
methods can be categorized into either optimization or filter based approaches.
In recent years, optimization based methods such as VINS-Mono \cite{Qin2018},
ORB-SLAM3 \cite{Campos2021orbslam3}, and OKVIS \cite{Leutenegger2015okvis}
have become
popular due to their ability to optimize a trajectory of poses. This is in
contrast with methods based on Extended Kalman
Filters (EKF) which marginalize all previous states. However, some modern
filtering methods have achieved
competitive performance via stochastic cloning \cite{Geneva2020}. For a more
detailed review, readers can refer to comparisons in \cite{Delmerico2018} and
\cite{He2020}.

MC-VIO has been less extensively studied because of hardware
complexity and potentially high computational requirements. Recent work from
\etalcite{Jaekel}{Jaekel2020} fused two pairs of stereo cameras and accounted
for the uncertainty in the extrinsics in both their front- and back-end systems
to improve performance. However, they had limited real-world experiments and no
ground truth comparisons.

A VIO system by \etalcite{Liu}{Liu2018}, with three pairs of stereo
cameras, estimated poses by minimising photometric errors. This work was
developed and tested in autonomous driving scenarios where the authors showed
the benefit of a multi-camera system when driving at night. However, since a
constant velocity motion model was used, their method is not suitable for
handheld sensing or robots with highly dynamic motions.

A unique omnidirectional setup was proposed by \etalcite{Seok}{Seok2020} with
four large FoV cameras. The four overlapping image regions were treated as
four stereo cameras, but the system did not fully take advantage of the
camera setup to track features across the camera pairs.

With a focus on aerial robotics, \etalcite{M\"uller}{Muller2018} used two pairs
of stereo cameras on an MAV (looking up and down). The images from their
wide-angle cameras were split in half and fed into four separate stereo VIO
systems running on a Field Programmable Gate Array (FPGA). The VIO outputs and
an IMU were then fused together using an EKF in a loosely coupled fashion.
However, the experimental results were limited to an indoor office building.

\etalcite{Kuo}{Kuo2020} introduced a more general design for multi-camera
Simultaneous Localization and Mapping (SLAM), which involved an adaptive
initialization scheme, keyframe
selection, and map management. Their system was based on SVO
\cite{Forster2017svo} and their approach required minimal
parameter tuning. However, their real-world experiments showed little
improvement in accuracy when using a multi-camera setup.

Meanwhile, using a Multi-State Constraint Kalman Filter (MSCKF),
\etalcite{Eckenhoff}{Eckenhoff2021} fused six cameras with multiple IMUs.
The MSCKF's low computational requirements allowed for asynchronous camera
keyframe processing and real-time operation. They presented detailed simulation
results, however, real-world experiments were limited to a small lab
environment.

We introduce a MC-VIO framework to fuse an arbitrary number of
stereo and monocular cameras, with a focus on multi-camera feature selection
and tracking.

We are particularly motivated to address common challenges and failure cases in
mono and stereo VIO, which have not been a focus of recent studies.
Hence, our
system is designed and tested in a variety of challenging environments, covering
fast and abrupt motions, severe illumination changes, indoor/outdoor scenes, and
dark underground environments.

\section{Problem Statement}
\label{sec:problem-statement}

\begin{figure}
\centering
\vspace{0.5mm}
\includegraphics[height=3.1cm]{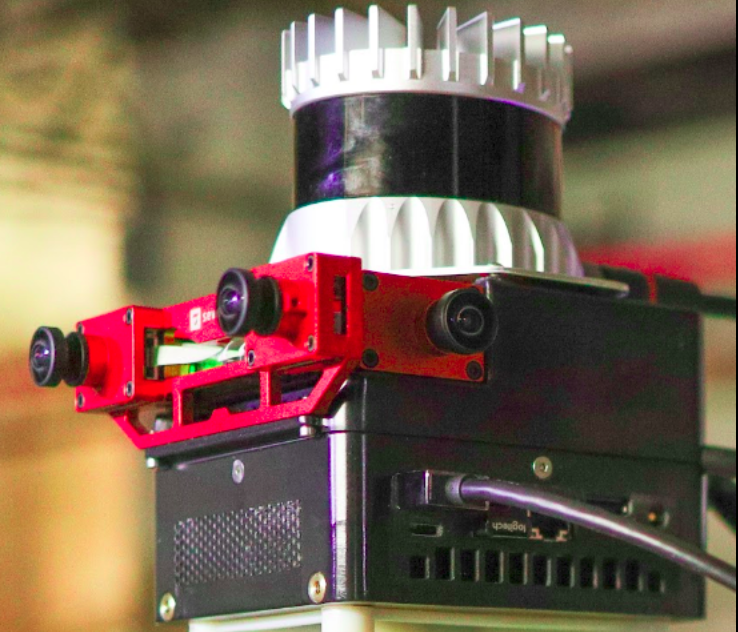}
\hspace{0.5mm}%
\includegraphics[height=3.1cm]{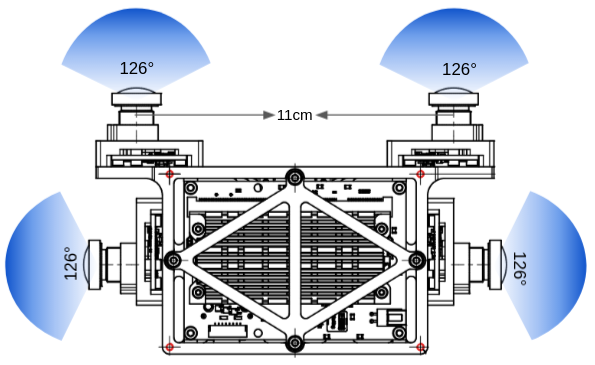}
\caption{VILENS-MC is evaluated using a custom built multi-camera handheld
device. See details in Section \ref{sec:hardware}.
}
\label{fig:halo-device}
\vspace{-2mm}
\end{figure}

We aim to estimate the position, orientation, and velocity of a mobile
platform equipped with an IMU and multiple hardware-synchronized cameras.
\Figure \ref{fig:halo-device} shows the multi-camera device used
in for the majority of the experimental results.

The relevant reference frames are as follows: the earth-fixed
world frame $\World$, the platform-fixed base frame $\Base$, the IMU frame
$\Imu$, and $n$ individual camera frames $\Camera_c$ where $c \in \{1, ... ,
n\}$.

Unless otherwise specified, the position
$\tensor[_\world]{\tran}{_{\world\base}}$ and orientation $\R_{\world\base}$
of the base (with $\tensor[_\world]{\T}{_{\world\base}} \in \SEthree$ as the
corresponding homogeneous transform) are expressed in world
coordinates; the linear velocity is in the base frame
$\tensor[_\base]{\vel}{_{\world\base}}$, and IMU biases
$\tensor[_\imu]{\bias}{^{g}},\;\tensor[_\imu]{\bias}{^{a}}$ are expressed in
the IMU frame.

\subsection{State and Measurements Definition}

\noindent The state of the sensor rig at time $t_i$ is defined as
follows,
\begin{equation}
\State_i \triangleq \left[\R_i,\tran_i,\vel_i, \bias^{g}_i \,\; \bias^{a}_i
\right] \in \SOthree \times \Real^{12}
\end{equation}
where: $\R_i$ is the orientation, $\tran_i$ is the position, $\vel_i$ is the
linear velocity, and $\bias^{g}_i$, $\bias^{a}_i$ are, respectively, the usual
IMU gyroscope and
accelerometer biases. In addition to the states, we track point landmarks
$\Landmark_{\ell}$ as
triangulated visual features.

The objective is to estimate the optimized trajectory $\X_k$ of all states
$\State_{i}$ and landmarks
$\Landmark_{\ell}$ visible up to the current time
$t_k$ within a fixed lag smoothing window.

The measurements from the set of multiple cameras $\calC$ and an IMU $\calI$ are
hardware synchronized but received at different frequencies. We define $\Z_k$ as
the full set of measurements received within the smoothing window.

\subsection{Maximum-a-Posteriori Estimation}
We maximize the likelihood  of the measurements $\calZ_k$,
given the history of states $\calX_k$,
\begin{equation}
 \X^*_k = \argmax_{\X_k} p(\X_k|\Z_k) \propto
p(\X_0)p(\Z_k|\X_k)
\label{eq:posterior}
\end{equation}
The measurements are formulated as conditionally independent and
corrupted by white Gaussian noise. Therefore, \Equation
\eqref{eq:posterior}
can be expressed as the following least squares minimization
\cite{Dellaert2017},
\begin{multline}
\X^{*}_k = \argmin_{\X_k} \|\mathbf{r}_0\|^2_{\Sigma_0} + \\
\sum_{i \in \mathsf{K}_k}   \left(
\|\mathbf{r}_{\calI_{ij}}\|^2_{\Sigma_{\calI_{ij}}}
+ \sum_{\ell \in \mathsf{M}_i} \|\mathbf{r}_{\State_i,\Landmark_{\ell}}
\|^2_{\Sigma_{\State_i, \Landmark_\ell}}
\right)
\label{eq:cost-function}
\end{multline}
where $\calI_{ij}$ are the IMU measurements between $t_{i}$ and
$t_j$ and $\mathsf{K}_k$ are all the keyframes indices in the sliding window
up to $t_k$.
Each term is the residual associated to a factor type, weighted by the
inverse of its covariance matrix. The residuals include prior, IMU, and visual
landmark factors.

\begin{figure}
 \centering
  \vspace{0.5mm}
 \includegraphics[width=0.9\columnwidth]{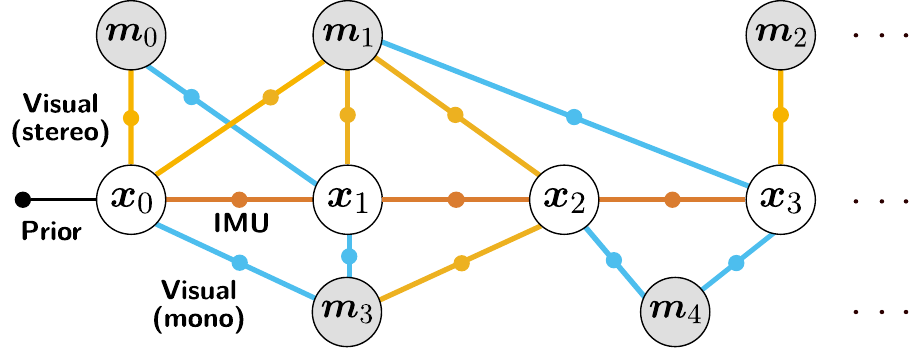}
 \caption{Sliding window factor graph structure, showing prior, visual,
 and preintegrated IMU factors. Landmarks can be tracked across both stereo and
 mono
 cameras to create longer feature tracks and improve accuracy.
 }
 \label{fig:factor-graph}
 \vspace{-3mm}
\end{figure}

\section{Factor Graph Formulation}
\label{sec:twist-factors}

In this section, we describe the measurements, residuals, and
covariances of the IMU and visual factors shown in \Figure
\ref{fig:factor-graph}.

\subsection{Preintegrated IMU Factors}
\label{sec:imu-factors}
We follow the standard manner of IMU measurement
preintegration from \cite{Forster2017a} to constrain the pose, velocity, and
biases
between two consecutive nodes of the graph, providing high frequency state
updates between nodes. The residual has the following form,
\begin{equation}
\mathbf{r}_{\calI_{ij}}  = \left[ \mathbf{r}^\transpose_{\Delta
\R_{ij}}, \mathbf{r}^\transpose_{\Delta \vel_{ij}},
\mathbf{r}^\transpose_{\Delta \tran_{ij}},
\mathbf{r}_{\bias^a_{ij}},
\mathbf{r}_{\bias^g_{ij}} \right]
\end{equation}
For a detailed definition of the above residuals, see \cite{Forster2017a}.

\subsection{Stereo and Mono Landmark Factors}
\label{sec:visual-factors}
We define a visual landmark in
Euclidean space as $\Landmark_\ell \in
\Real^3$. Given the platform pose
$\T_i$ (for
simplicity, we omit the fixed transform between base and camera), we can
project $\Landmark_\ell$ onto the image plane with the function
 $\pi: \SEthree \times \Real^3 \mapsto \Real^2$, resulting in the projected
coordinates $(u_{\ell}, v_{\ell}) \in
\Real^2$ on the image plane (orange/green circles in \Figure
\ref{fig:cross-cam-tracking}).  Thus, the residual at state $\State_i$ for
landmark
$\Landmark_\ell$ is defined as \cite{Wisth2021},
\begin{equation}
	\mathbf{r}_{\State_i, \Landmark_\ell} =
	\left( \begin{array}{c}
		\pi_u^L(\T_i, \Landmark_\ell) - u^L_{i,\ell} \\
		\pi_u^R(\T_i, \Landmark_\ell) - u^R_{i,\ell} \\
		\pi_v(\T_i, \Landmark_\ell) - v_{i,\ell}
	\end{array} \right) \label{eq:stereo-residual}
\end{equation}
where $(u^{L}, v), (u^{R}, v)$ are the pixel locations of
the detected landmark. $\Sigma_{\Landmark}$ is computed using an uncertainty
of \SI{0.25}{pixels}. We also account for lens distortion using the covariance
warping method from \cite{Wisth2021tro}. For
monocular landmarks, only the first and the last rows of \Equation
\eqref{eq:stereo-residual} are used.

The location of the landmarks detected by the stereo camera pair is
initialized using stereo triangulation. For landmarks detected in monocular
cameras, we triangulate the feature location over the last $N_{obs}$ frames
using the Direct Linear Transform (DLT) algorithm from
\cite{Hartley2004dlt}.


\section{Cross Camera Feature Tracking}
\label{sec:feature-tracking}

\begin{figure}
\centering
\includegraphics[width=\columnwidth, trim={0.25cm 0 0.25cm 0},
clip]{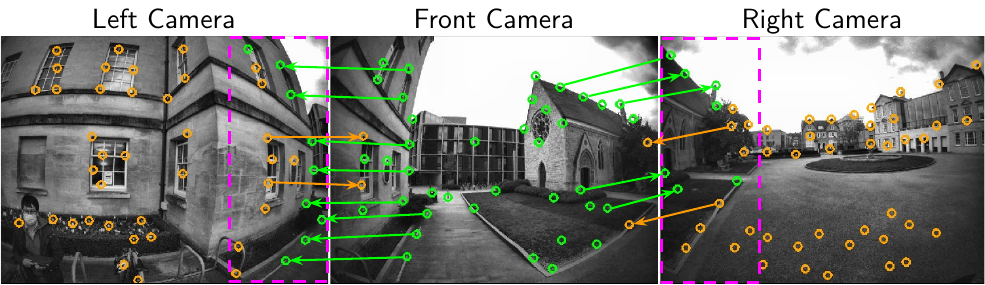}
\caption{VILENS-MC takes advantage of any overlapping image regions (purple
rectangles) in a
multi-camera setup to track features across cameras. This increases feature
track length and avoids tracking the same feature independently in different
cameras. The arrows indicate features being tracked from one image to another.}
\label{fig:cross-cam-tracking}
\vspace{-1mm}
\end{figure}

Most multi-camera systems do not take advantage of the overlapping FoVs of
their setups, which produce common image regions that allow for cross-camera
feature matching.

We propose a simple and effective method to continuously track features across
different cameras which we call Cross Camera Feature Tracking (CCFT). CCFT can
avoid the optimization of redundant landmarks by tracking the same feature in
different cameras simultaneously. As shown in \Figure \ref{fig:factor-graph},
we continually add constraints between states $\State_i$ and landmarks
$\Landmark_\ell$ even as they are tracked across cameras. For example, landmark
$\Landmark_1$ is first tracked in a stereo camera at $\State_0, \State_1,
\State_2$, then tracked in a monocular camera at $\State_3$.

A common approach to perform this feature matching would be via feature
descriptors. However, this would be computationally expensive for real-time,
high-frequency VIO on a mobile platform. For example, the combination of
feature extraction and stereo matching in ORB-SLAM2
\cite{Mur-Artal2017} takes $\sim$\SI{24.8}{\milli\second} which is
significant for a \SI{30}{\hertz} image stream. Instead, we make use of the
known camera extrinsics and the preintegrated IMU measurements between states
to associate features geometrically.

When feature depth is available at the current image time $t_k$
(\eg from stereo camera triangulation), then we can directly project the visual
landmark $\Landmark_\ell$ from camera $\Camera_i$ into camera $\Camera_j$ via
the extrinsic
transformation between cameras $\T_{\Camera_i\Camera_j}$,
\begin{equation}
(u, v)_{\ell, c_j, t_k} = \pi_{c_j}\left(\T_{\Camera_j\Camera_i}\;
\tensor[_{\Camera_i}]{\Landmark}{_{\ell, t_k}}\right)
\end{equation}
Alternatively, if the feature depth is not directly available at the current
timestamp (\eg if the feature is only tracked in a monocular camera), then we
instead use the preintegrated IMU measurements to estimate the landmark
location at the current time,
\begin{equation}
(u, v)_{\ell, c_j, t_k} = \pi_{c_j} \left(\T_{\Camera_j\Camera_i} \;
\hat{\T}_{t_{k},t_{k-1}}\; \tensor[_{\Camera_i}]{\Landmark}{_{\ell,
t_{k-1}}}\right)
\end{equation}
where $\hat{\T}_{t_{k},t_{k-1}}$ is the estimated transform between the
previous camera pose at $t_{k-1}$ and the current time $t_k$.

This process is completed for each feature in the overlapping image regions.
The projected feature location is refined by matching it to the closest
image feature in camera $c_j$, using a Euclidean distance metric. This is a
highly effective method in practice, assuming there is a good extrinsic
calibration. Any incorrect associations are handled by the optimizer using
robust cost functions.

\Figure \ref{fig:cross-cam-tracking} shows an example of the feature matches
using CCFT. The purple areas highlight
the overlapping image regions where features can be tracked across cameras. The
CCFT method typically reduces the number of landmarks added into the
optimization back-end and gives improved estimation accuracy. This is discussed
further in Section \ref{sec:result-cross-cam}.

\section{Submatrix Feature Selection}

In general, increasing the number of features tracked in a VIO system
improves the estimation accuracy \cite{Liu2018}. However, it also
increases
computation, eventually reaching a point where the algorithm can fail due to
computational constraints. This is a particular problem in multi-camera
systems, where more features can be tracked than the optimizer can handle. Thus,
it is important to track and optimize only the best features.

Specifically, our SFS algorithm is based on
\cite{Zhao2020}, where the authors presented a feature selection algorithm to
reduce the computational cost of active map-to-frame feature matching.
We instead apply this algorithm to MC-VIO and implement changes to improve
numerical stability.
The aim is to maintain accuracy while reducing the optimization time of MC-VIO.

In this section, we first outline the construction of the joint feature
Jacobian and
covariance matrices. Then, we describe the \textit{Max-logDet} algorithm
to
select the most representative subset of tracked features to be optimized.

\subsection{Construct Joint Feature Jacobian Matrix}

The optimization aims to minimize the cost function defined in
\Equation \eqref{eq:cost-function}. Since we are interested in selecting the
best features in each frame $k$, we focus on improving the conditioning of the
landmark residual, where $\Measurement_{\ell}$ is the feature location of $(u,
v)$,
\begin{align}
\X^* &= \argmin_{\X} \sum_{\ell \in \mathsf{M}} \|
\mathbf{r}_{\State,\Landmark_{\ell}}
\|^2_{\Sigma_{\Landmark_\ell}}
\\
&= \argmin_{\X} \sum_{\ell \in \mathsf{M}}  \|
\pi(\T, \Landmark_\ell) - \Measurement_{\ell}
\|^2_{\Sigma_{\Landmark_\ell}}
\end{align}
Stacking $\Landmark_\ell$, $\Measurement_\ell$ into column matrices
$\Landmarks$, $\Measurements$ we get,
\begin{equation}
\X^* = \argmin_{\X} \| \Pi(\T, \Landmarks) - \Measurements \|^2_{\Sigma}
\end{equation}
Applying the method from \cite{Zhao2020}, we perform a
first order linearization around the
initial value $\T_g$,
\begin{equation}
\| \Pi(\T, \Landmarks) - \Measurements \|^2_{\Sigma} \approx
\| \Pi(\T_g, \Landmarks) + \hjac_\T(\T \ominus \T_g) - \Measurements
\|^2_{\Sigma}
\label{eq:jac-cost-function}
\end{equation}
where $\hjac_\T$ is the pose Jacobian linearized about the value
$\T_g$, and
$\ominus$ is the Lie group difference operator between poses
\cite{Wisth2021tro}.

To minimize \Equation \eqref{eq:jac-cost-function}, we can use Gauss-Newton
optimization to iteratively update the pose estimate,

\begin{equation}
\T_{g+1} = \T_g - \hjac_\T^+ \left(\Pi(\T_g, \Landmarks) -
\Measurements \right)
\label{eq:jac-cost-linearization}
\end{equation}
where $\hjac_\T^+$ is the Moore-Penrose pseudo-inverse of $\hjac_\T$. This
method has
converged (\ie reached minimal error) when the update step is approximately
zero, \ie
\begin{equation}
\boldsymbol{\epsilon_\T} = \hjac_\T^+ \left( \Pi(\T_g, \Landmarks) -
\Measurements
\right)
\approx 0
\end{equation}
Performing another first order linearization around $\Landmarks_g$,
\begin{align}
\boldsymbol{\epsilon_\T} &\approx \hjac_\T^+ \left(\Pi (\T_g, \Landmarks_g) -
\Measurements + \hjac_{\Landmarks} (\Landmarks - \Landmarks_g)
\right)\\
\boldsymbol{\epsilon_\T} &=  \hjac_\T^+ \left (\boldsymbol{\epsilon_Z} -
\hjac_{\Landmarks} \boldsymbol{\epsilon_{\Landmarks}}
\right)
\end{align}
where $\hjac_{\Landmarks}$ is the landmark-to-image projection
Jacobian. We
define the measurement error as $\boldsymbol{{\epsilon}_Z} = \Pi (\T_g,
\Landmarks_g)
-
\Measurements$, and
the landmark error as $\boldsymbol{\epsilon_{\Landmarks}} =
\Landmarks - \Landmarks_g$.

Without loss of generality, the measurement error can be modeled as
$\boldsymbol{\epsilon_{z_\ell}}\sim \calN(0, \Sigma_{z_\ell})$ which is
influenced by image processing parameters and noise.
However, if biased landmark errors exist, then the mean of
error distribution is non-zero. This could be caused by batch perturbation of
the landmark positions by incorrect camera poses, or a slight scale error.
Hence the landmark error is modeled as non-zero-mean i.i.d. Gaussian,
$\boldsymbol{\epsilon_{\Landmark_\ell}} \sim \calN(\mu_{\Landmark_\ell},
\Sigma_{\Landmark_\ell})$, such that the expected value of
$\boldsymbol{\epsilon_\T}$ is,
\begin{align}
{\mathbb{E}}[\boldsymbol{\epsilon_\T}]
&= \mathbb{E} \left[\hjac_\T^+\left(\boldsymbol{\epsilon_Z} -
\hjac_{\Landmarks} \boldsymbol{\epsilon_{\Landmarks}}\right) \right] \\
{\mathbb{E}}[\boldsymbol{\epsilon_\T}]  &= \hjac_\T^+
\hjac_{\Landmarks}{\textbf{1}_n}\mu_{\Landmarks} \\
\hjac_{\Landmarks}^ + \hjac_\T\mathbb{E}[\boldsymbol{\epsilon_\T}] &=
{\textbf{1}_n}\mu_{\Landmarks}
\end{align}
where $\textbf{1}_n$ is a tall matrix made of identity matrices to transform the
dimension of $\mu_{\Landmarks}$. $\hjac_{\Landmarks}$ consists of $n$
diagonal
blocks made up of individual landmark
Jacobians $\hjac_{\Landmark_\ell} \in\Real^{2 \times 3}$, where $n$ is
the number of landmarks with known 3D
positions. $\hjac_\T$ consists of pose Jacobians for each landmark
$\hjac_{\T, \Landmark_\ell} \in\Real^{2 \times 6}$.

To obtain a joint Jacobian matrix $\hjac_J$, we add an additional row
of zeros to
$\hjac_{\T, \Landmark_\ell}$ and a row of $[0, 0, 1]$ to
$\hjac_{\Landmark_\ell}$. This
makes $\hjac_{\T, \Landmark_\ell}$ invertible without changing the
structure of the least squares problem \cite{Zhao2019}.
Performing block-wise multiplication results in the matrix $\hjac_J$,
\begin{equation}
\hjac_J = \begin{bmatrix}
\hjac_{\Landmark_0}^{-1} \hjac_{\T, \Landmark_0}& 
... & 
\hjac_{\Landmark_{n-1}}^{-1} \hjac_{\T_, \Landmark_{n-1}}
\end{bmatrix}^\transpose
\end{equation}
from which the simplified pose covariance matrix follows,
\begin{equation}
\Sigma_\T = \hjac_J^+ (\hjac_J^+)^\transpose =
(\hjac_J^\transpose
\hjac_J)^{-1}
\end{equation}

\subsection{Submatrix Selection}

Since the overall pose error depends on the spectral properties of $\hjac_J$, we
aim to find a submatrix $\hjac_{sub}$ (\ie subset of features) that best
preserves this distribution. This can be understood as finding row blocks in
$\hjac_J$
to maximize the norm of the selected submatrix.

We adapt the stochastic sampling method presented in \cite{Zhao2020} for
feature selection. We choose the \textit{Max-logdet} metric since it best
approximates the original full feature set in visual
odometry \cite{Zhao2019, Carlone2019}. Additionally, the combination
of greedy (deterministic) and stochastic sampling (randomized acceleration)
gives the best approximation ratio of any polynomial time algorithm
\cite{Mirzasoleiman2015}.

We note that since $\hjac_{J_i} \in \Real^{2 \times 6}$ is not a full rank
matrix,
the determinant of its square $\hjac_{J_i}^\transpose \hjac_{J_i}$ would be
zero. In
contrast to \cite{Zhao2020}, we add a small diagonal matrix $\lambda \Identity
\in \Real^{6\times 6}$ to the
$\log\det$ calculation,
\begin{equation}
\hjac_{J_i} = \mathbf{\arg\max}\left( \log \det
(\hjac_{J_i}^\transpose \hjac_{J_i} + \hjac_{sub}^\transpose \hjac_{sub} +
\lambda \Identity) \right)
\end{equation}
This makes the squared matrix full rank with a
non-zero determinant while preserving its spectral properties. An alternative
method would be projecting $\hjac_{J_i}^\transpose \hjac_{J_i}$ onto a full
rank subspace, which we will consider in future work.

We apply this algorithm to each camera individually in our multi-camera setup.
An ablation study showing how this method maintains high accuracy and improves
optimization time is presented in Section
\ref{sec:ablation-study-inliear-selection}.

\section{Implementation}
\label{sec:implementation}

\begin{figure}
	\centering
	\vspace{0.5mm}
	\includegraphics[width=\columnwidth]{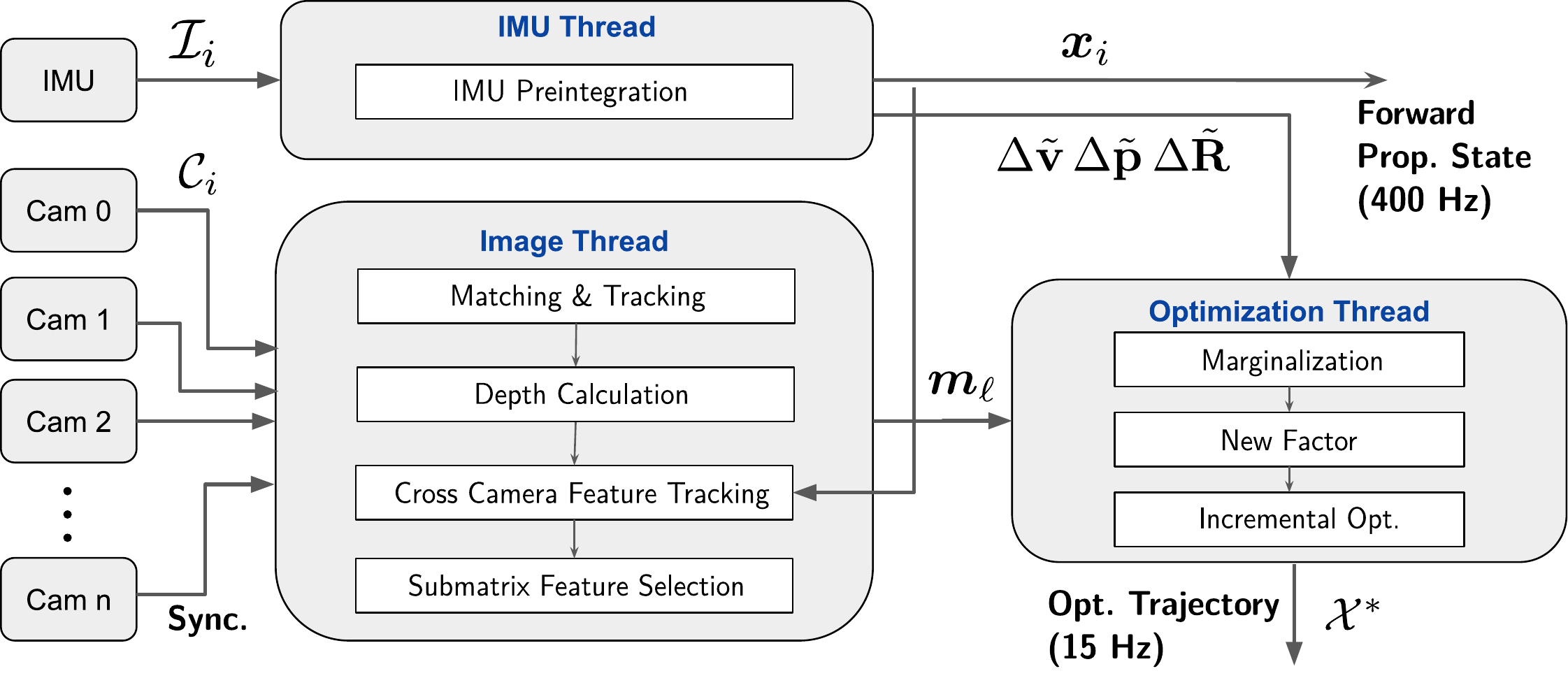}
	\caption{Overview of the VILENS-MC architecture. The IMU and cameras
	are handled in separate threads by front-end measurement handlers.
	The back-end produces both a high frequency forward-propagated output and a
	lower frequency optimized output.}
	\label{fig:state-estimation-architecture}
	\vspace{-1mm}
\end{figure}

An overview of the system architecture is shown in \Figure
\ref{fig:state-estimation-architecture}. Three parallel threads perform IMU
preintegration, camera image processing, and optimization, respectively. The
system outputs the optimized state estimate from the factor graph at camera
keyframe frequency (typically
\SI{15}{\hertz}) for navigation or mapping, while a forward-propagated state
at IMU frequency is also produced at \SI{400}{\hertz} for high
frequency tasks.

The factor graph is solved using fixed lag smoothing based on the
efficient incremental optimization solver iSAM2 from the
GTSAM library \cite{Dellaert2017}. For all of our experiments, we
used a lag time of \SI{3.5}{\second}. To reduce the effect of outliers, the
visual factors are added to the graph using the Dynamic Covariance Scaling (DCS)
robust cost function \cite{MacTavish2015}.

\subsection{Hardware and Calibration}
\label{sec:hardware}
The sensor used for our experiments is the Alphasense multi-camera development
kit from Sevensense Robotics AG, shown in \Figure \ref{fig:halo-device}. An
onboard FPGA synchronizes the IMU and four grayscale fisheye cameras -- a
frontal stereo pair with an \SI{11}{\centi\meters} baseline and two lateral
cameras. Each camera has a FoV of \SI{126 x 92.4}{\degree} and a resolution of
\SI{720 x 540}{\pixel}. This configuration produced an overlapping FoV
between the front and side cameras of about \SI{36}{\degree}. The cameras and
the embedded cellphone-grade IMU were operated at \SI{30}{\hertz} and
\SI{200}{\hertz}, respectively. Camera-IMU extrinsic and intrinsic
calibration was conducted offline using the Kalibr \cite{Rehder2016kalibr}
toolbox. The Ouster OS0-128 lidar shown in the figure was used only for
ground truth generation and all sensors were hardware synchronized with an Intel
NUC via Precision Time Protocol (PTP).

\subsection{Initialization}
We initialize the IMU biases by averaging the first \SI{1}{\second} of data at
system start up (assuming the IMU is stationary).

To solve the scale initialization problem, which is often present in monocular
visual odometry systems, we combine preintegrated IMU measurements
and depth from the stereo camera pair. Notably, the CCFT method allows features
from the stereo camera to flow into the monocular camera, speeding up the depth
initialization process.

\subsection{Visual Feature Tracking}
To get an even feature distribution across each image, we split them into $3
\times 3$ tiles before applying the FAST feature detector to each segment.
These features are tracked between frames using the KLT feature tracker and
across cameras using the CCFT method described above. Outliers are rejected
using RANSAC-based methods. Thanks to the incremental optimization and
multi-threading, every second frame
is used as a keyframe, achieving a \SI{15}{\hertz} nominal output.


\section{Experimental Results}
\label{sec:results}

We evaluated our algorithm on a variety of challenging indoor and outdoor
datasets, varying from narrow corridors to large open spaces. We compared our
algorithm to state-of-the-art methods, evaluating both quantitative
and qualitative performance.
We also include ablation studies showing how our proposed contributions reduced
Relative Pose Error (RPE) and decreased computation time. Finally, we
demonstrated the versatility of
this approach by applying VILENS-MC to a quadruped robot operating in an unlit
mine.

\subsection{Datasets}

\begin{figure}
\centering
\vspace{0.5mm}
\includegraphics[width=0.48\columnwidth]{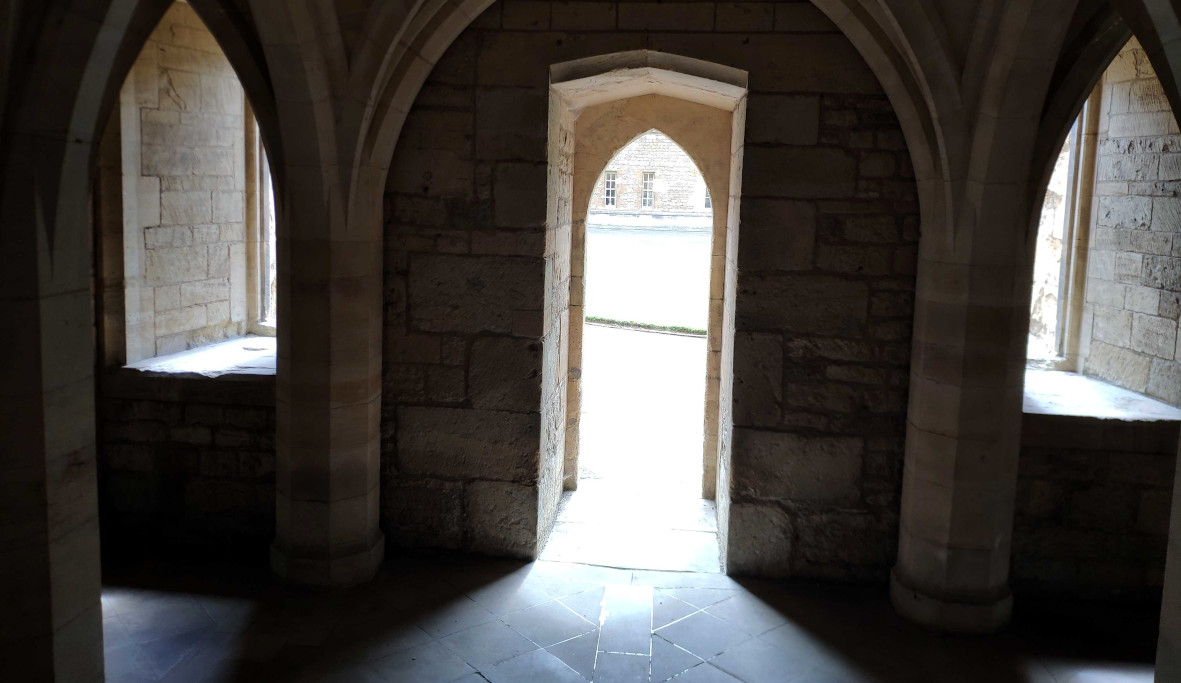}%
\hspace{1mm}%
\includegraphics[width=0.48\columnwidth]{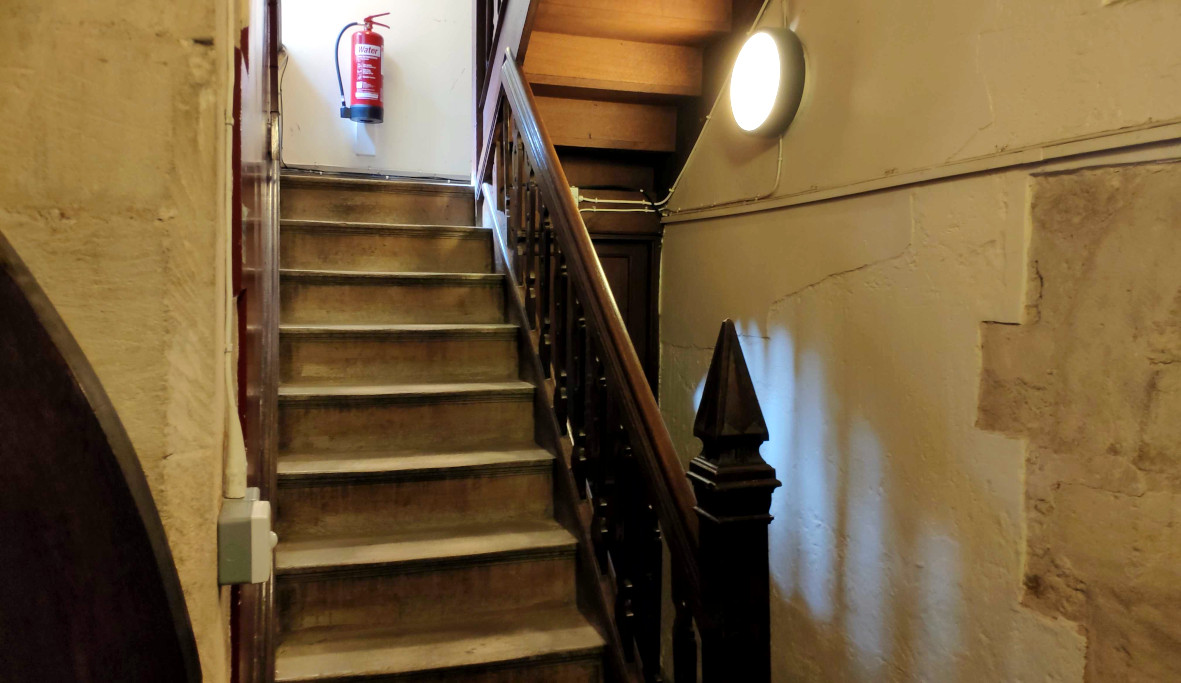} \\
\vspace{1mm}%
\includegraphics[width=0.48\columnwidth]{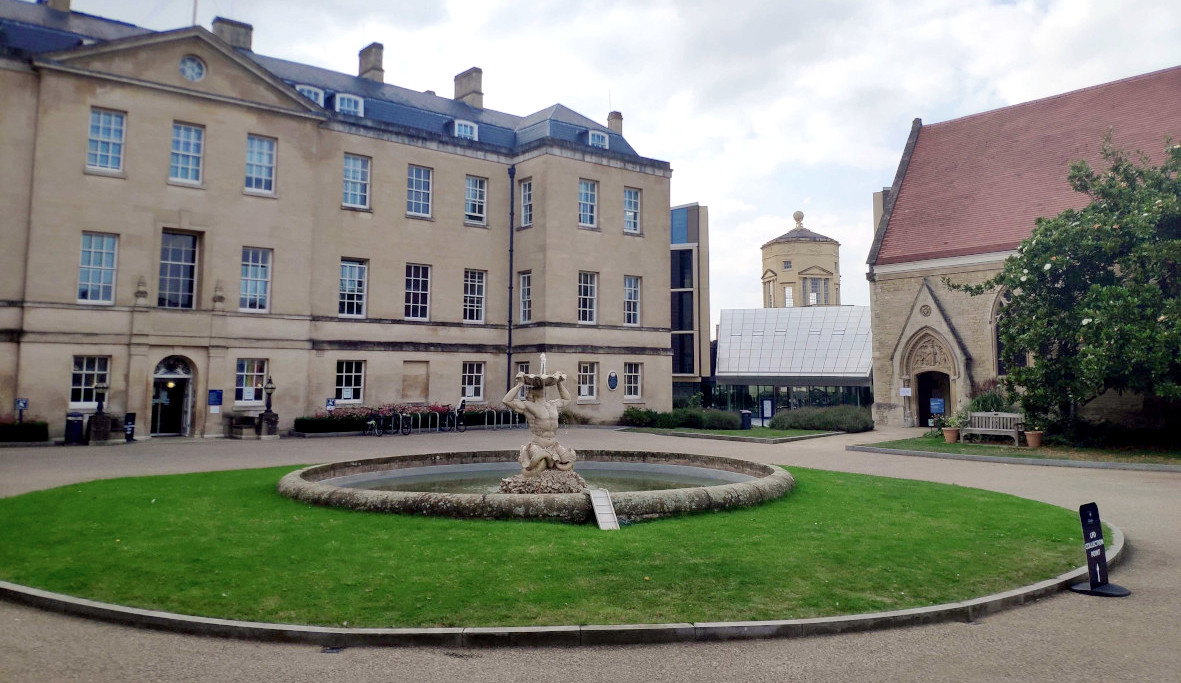}%
\hspace{1mm}%
\includegraphics[width=0.48\columnwidth]{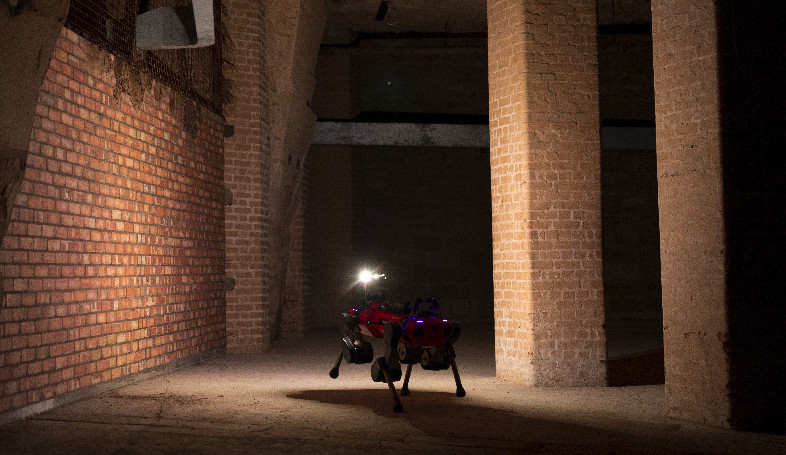}
\caption{ Test datasets: \textit{Top:} Newer College Dataset
\cite{zhang2021multicamera}.
\textit{Bottom Left:} Oxford Mathematical Institute. \textit{Bottom
Right:} A dark underground limestone mine.}
\label{fig:experimental-scenarios}
\end{figure}

\Figure \ref{fig:experimental-scenarios} gives an overview of the datasets used
for evaluation. These include additional multi-camera experiments collected for
the Newer College Dataset (NCD) \cite{zhang2021multicamera}, operation in an
unlit
underground limestone mine (MINE), and a large circuit around
the Oxford Mathematical Institute (MATH).

\begin{figure}
\centering
\includegraphics[width=\columnwidth]{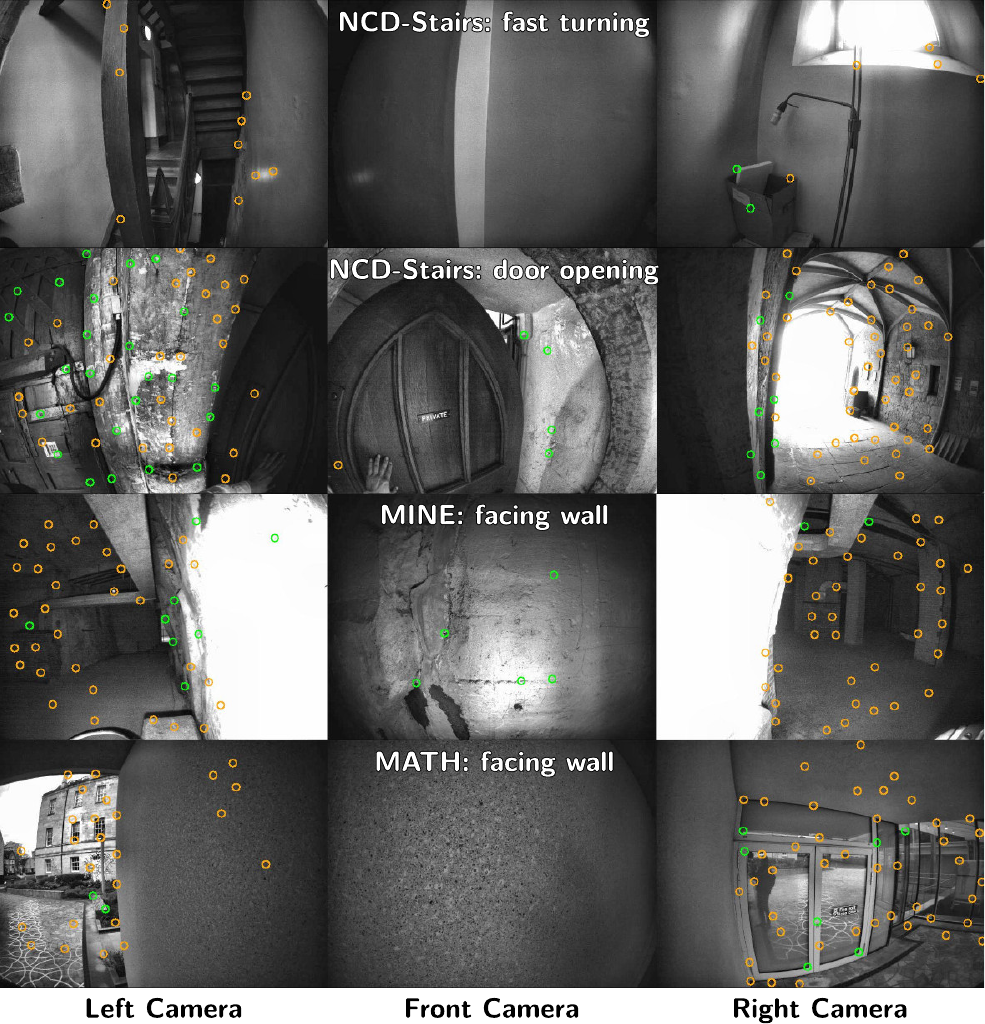}%
\caption{Examples of challenging scenes where it was difficult to track
features. Orange and green circles are features tracked by monocular
and stereo cameras, respectively (some have been tracked across images). Note
that stereo features were mostly absent or rejected in these examples.}
\label{fig:challenging-scenes}
\vspace{-2mm}
\end{figure}

These datasets were collected using the handheld multi-camera device described
in Section \ref{sec:hardware}. They were specifically chosen to test the limits
of VIO systems and include challenges such as aggressive shaking of the device
(up to \SI{5.5}{\radian\per\second}),
severe illumination changes, and loss of visual feature tracking in one or more
cameras (see \Figure \ref{fig:challenging-scenes}). All datasets were collected
at a fast walking pace of $\sim$\SI{1.5}{\meter\per\second}.
%
%
The specific experiments are:
\begin{itemize}
	\item \textbf{NCD-Quad:} Moving between narrow corridors and a large open
	college quadrangle, with strong illumination changes caused by bright
	sunlight (\SI{244}{\meter}, \SI{3}{\minute}).
	\item \textbf{NCD-Stairs:} Traversing a very narrow staircase, including a
	sequence where a door is opened directly in front of the camera
(\SI{59}{\meter},
	\SI{2}{\minute}).
	\item \textbf{MINE:} A medium scale, dark underground environment with
	regular illumination changes due to onboard lighting (\SI{236}{\meter},
	\SI{3}{\minute}).
	\item \textbf{MATH:} A large scale outdoor environment with aggressive
	shaking of the device up to \SI{5.5}{\radian/\second} (\SI{329}{\meter},
	\SI{4}{\minute}).
\end{itemize}
The multi-camera experiments which extend NCD (NCD-Quad, NCD-Stairs) have been
publicly released. \footnote{Available at
\url{https://ori-drs.github.io/newer-college-dataset/}.}

To generate the ground truth, Iterative Closest Point (ICP) registration was
used to align the current lidar scan to detailed prior maps collected from a
commercial laser scanner. The high frequency motion
estimate from the IMU was used to carefully remove
lidar motion distortion \cite{Wisth2021}. For an in-depth discussion on ground
truth generation, refer to \cite{ramezani2020newer}.

\subsection{Quantitative Analysis}
\begin{figure}
\centering
\vspace{0.5mm}
\includegraphics[width=1.0\columnwidth]{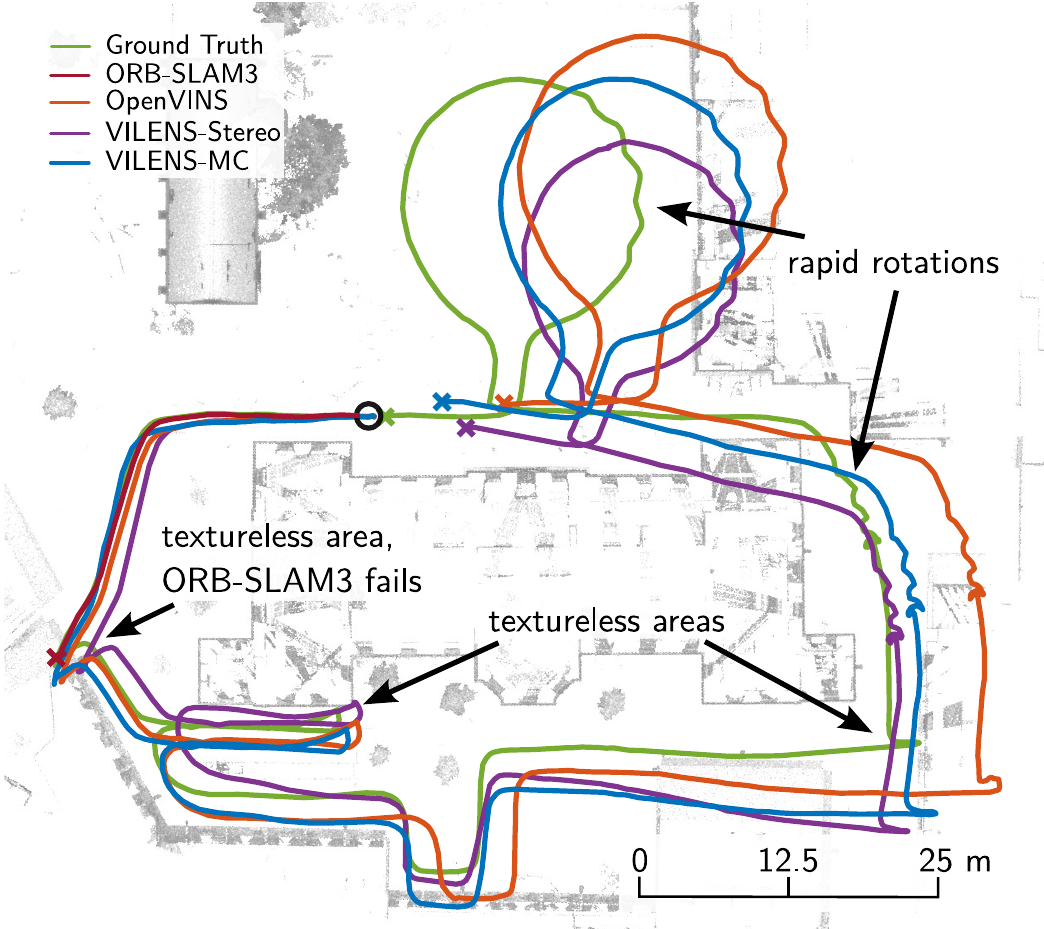}\\
\caption{Top-down view of \textbf{MATH} dataset comparing the estimated
trajectory of ORB-SLAM3, OpenVINS, and
VILENS-MC against the ground truth. A black circle
marks the start of the trajectory, while colored crosses indicate the last pose
of each trajectory. Textureless locations, where cameras were facing walls, are
shown in \Figure \ref{fig:challenging-scenes}, row 4.}
\label{fig:math-trajectory}
\end{figure}

\begin{table}
\centering
\resizebox{\columnwidth}{!}{%
\begin{tabular}{lcccc}
\toprule
\multicolumn{5}{c}{\textbf{ \SI{10}{\meter} Relative Pose Error (RPE) --
Translation
[\si{\metre}] / Rotation [\si{\degree}]}} \\
\midrule
 & \textbf{OpenVINS}
 & \textbf{ORB-SLAM3}
 & \textbf{VILENS-Stereo}
 & \textbf{VILENS-MC} \\
\midrule
Cameras  & 2 (Stereo) & 2 (Stereo) & 2 (Stereo) & 4\\
\midrule
NCD-QUAD & 1.01 / 1.26 & \textbf{0.23} / 0.91 & 0.30 / 1.24 & 0.31 /
\textbf{0.82}\\
NCD-STAIRS & 0.33 / 3.05 & 0.20* / 3.04 & 0.43 / 5.85 & \textbf{0.16} /
\textbf{2.56} \\
MINE & 0.98* / 2.10 & Fail & 0.41* / 3.57 & \textbf{0.20} / \textbf{1.65} \\
MATH & 0.65* / 1.69 & Fail & 0.56* / 1.57 & \textbf{0.26} / \textbf{1.03} \\
\bottomrule
\end{tabular}
} 
\vspace{1mm}
\caption{Performance Comparison Between Different Algorithms
\\ * = one or more instabilities occurred
}
\label{tab:multi-cam-rpe}
\vspace{-3mm}
\end{table}

Table \ref{tab:multi-cam-rpe} shows the RPE over
\SI{10}{\metre} for OpenVINS \cite{Geneva2020}, ORB-SLAM3
\cite{Campos2021orbslam3} and VILENS \cite{Wisth2021tro} (which our system
builds upon). We refer to the configuration of VILENS which uses only stereo
and IMU input as \textit{VILENS-Stereo}.
Note we were not aware of any open-source MC-VIO algorithm for
comparison.

Compared to our baseline (VILENS-Stereo), we reduced mean RPE by
\SI{45}{\percent} / \SI{50}{\percent} in translation/rotation, while
tracking fewer features due to SFS. VILENS-MC also outperformed
other state-of-the-art stereo VIO
algorithms by up to \SI{80}{\percent} / \SI{39}{\percent} in
translation/rotation.

In benign VIO conditions, such as those found in the NCD-Quad dataset, the
performance of VILENS-MC was similar to state-of-the-art systems. In
particular, ORB-SLAM3 appeared to outperform the other algorithms by
taking advantage of its local map matching SLAM system to correct for failure
events when the camera revisited a previous location.

However, in the MINE and MATH datasets, where there were few or
no stereo features, VILENS-MC performed well, even when the other stereo VIO
systems failed or had significant drift. This highlights a key
benefit of the tight fusion of multiple cameras, the natural robustness to
scene degradation.

\begin{figure}
\centering
\vspace{0.5mm}
\includegraphics[width=\columnwidth]{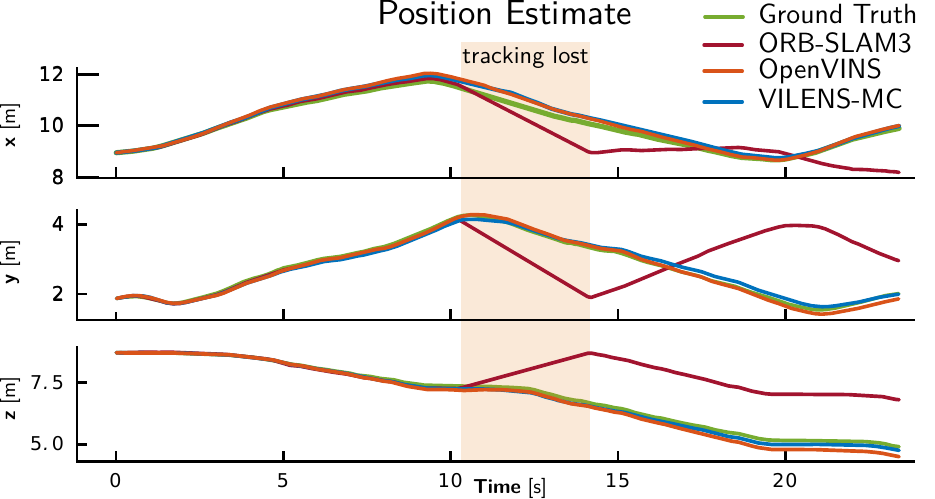}
\caption{ORB-SLAM3 lost tracking and failed to estimate odometry for
\SI{4}{\second}
in NCD-Stairs (scenario shown in \Figure \ref{fig:challenging-scenes} Row 3).
Note that elevation ($z$ axis) is properly tracked afterwards.}
\label{fig:orbslam-lost}
\end{figure}

\subsection{Degeneracy and Failure Analysis}
\subsubsection{MATH}
A top-down view of the trajectories in the MATH dataset is shown in \Figure
\ref{fig:math-trajectory}. In challenging scenarios where stereo camera feature
tracking failed (marked as ``textureless areas''), stereo VIO algorithms
continued to operate but relied on the IMU only,
causing the RPE to quickly increase, while VILENS-MC maintained robust
estimation by utilizing multiple cameras looking in different directions.

\subsubsection{NCD-Stairs}
\Figure \ref{fig:orbslam-lost} shows an extract from the
NCD-Stair dataset. Note how ORB-SLAM3 suffered from significant drift when
no features were detected in the stereo camera for \SI{4}{\second} until it
re-initialized (orange area). Even though the average RPE over the entire
NCD-Stair sequence remained low, a gap in estimations would be undesirable for
an autonomous
robot.

\subsubsection{MINE}
\begin{figure}
\centering
\includegraphics[width=\columnwidth]{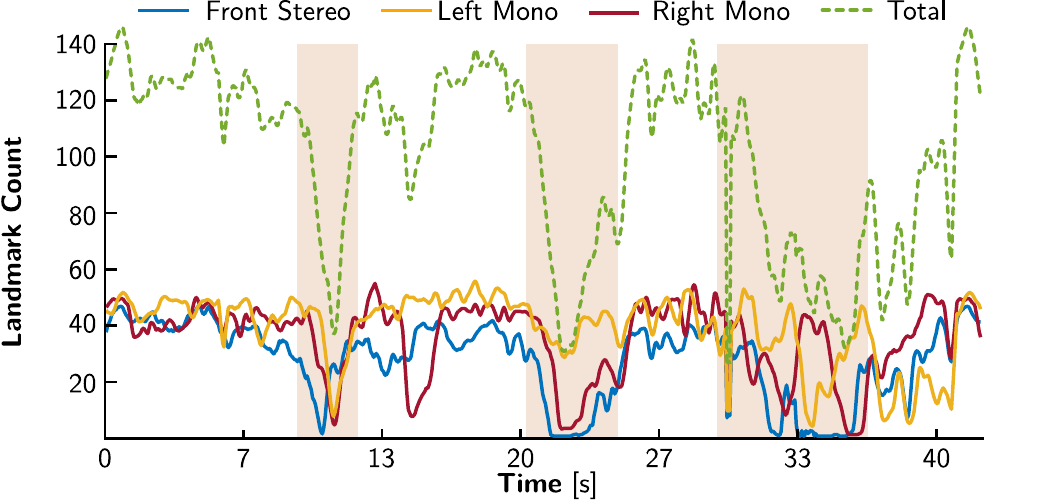}
\caption{A \SI{40}{\second} section from the \textbf{MINE}
dataset showing landmark count per each camera: front stereo pair (solid blue),
left mono camera (solid yellow), right mono camera (solid red). The total number
of landmarks is dashed green.}
\label{fig:corsham-lose-features}
\end{figure}

\Figure \ref{fig:corsham-lose-features}
shows the number of tracked landmarks during a \SI{40}{\second} interval
in the MINE dataset. During this time, several instances occurred where feature
tracking failed in either one or two cameras. Failures occurred when the device
entered a dark room ($\sim$\SI{10}{\second}), when it approached a
wall and turned on the spot ($\sim$\SI{20}{\second}, shown in \Figure
\ref{fig:challenging-scenes} row 3), and when it entered a dark
corner where only one camera could observe features at a time
($\sim$\SI{30}{\second}). By tracking features across multiple cameras,
VILENS-MC maintains accurate estimation throughout these events.

\section{Ablation Study}
\subsection{Cross Camera Feature Tracking}
\label{sec:result-cross-cam}
In this section, we evaluate the benefit of the CCFT method (Section
\ref{sec:feature-tracking}). CCFT enables continuous
feature tracking over longer periods and adds additional constraints to
landmarks.

Table \ref{tab:ablation-study} summarizes the effect this
method had on mean RPE (over \SI{10}{\meters}), showing that on average CCFT
decreased both translational and rotational RPE on all datasets except for
translation in NCD-QUAD. We attribute this to the system's susceptibility to
extrinsic multi-camera calibration errors.

Additionally, CCFT can reduce the
number of landmarks in the optimization by $\sim$\SI{10}{\percent}. Since our
hardware setup (ref. Section \ref{sec:hardware}) had only a small overlapping
FoV between frontal and lateral cameras, we expect to see a greater benefit with
a larger FoV overlap.

\subsection{Submatrix Feature Selection}
\label{sec:ablation-study-inliear-selection}

\begin{table}
	\centering
	\vspace{2mm} 
\resizebox{\columnwidth}{!}{%
	\begin{tabular}{l|cc||ccc}  \toprule
	& \multicolumn{5}{c}{\textbf{Relative Pose Error -- Translation
[\si{\metre}] / Rotation [\si{\degree}]}} \\
		\midrule
		& \multirow{2}{*}{\textbf{Baseline}} &
		\multirow{2}{*}{\textbf{CCFT}} &
		\textbf{Baseline}& \textbf{Baseline} & \textbf{SFS}\\
		& & & (90 feat.) & (150 feat.) & (90 feat.)\\
		\midrule
		NCD-QUAD & \textbf{0.27} / \textbf{0.77} & 0.34 / 0.88 & 0.37 / 1.02 &
		0.34 / 0.88 & \textbf{0.31} / \textbf{0.82} \\
		NCD-STAIR & 0.17 / 3.60 & \textbf{0.16} / \textbf{2.85} & 0.28 / 5.06 &
		\textbf{0.16} / \textbf{2.85} & 0.18 / 3.03\\
		MINE & 0.25 / 1.91 & \textbf{0.21} / \textbf{1.64} & 0.34 / 2.70 & 0.21
		/ \textbf{1.71} & \textbf{0.20} / 1.75\\
		MATH & 0.42 / 1.01 & \textbf{0.28} / \textbf{1.01} & 0.33 / 1.16 & 0.28
		/ \textbf{1.01} & \textbf{0.26} / 1.03\\
		\bottomrule
	\end{tabular}
	} 
	\vspace{1mm}
\caption{Ablation Studies: Cross Camera Feature Tracking (CCFT) and Submatrix
Feature Selection (SFS)}
\label{tab:ablation-study}
\vspace{-6mm}
\end{table}

This ablation study demonstrates the benefit of SFS. The baseline
approach simply spreads features evenly across the image based on their FAST
score and mutual pixel distance.

A comparison between the baseline and proposed SFS methods is
shown in Table \ref{tab:ablation-study}. The first two columns show the baseline
with 90 and 150 features tracked and optimized across all cameras,
while the SFS method selected only 90 out of 150 tracked features to optimize.

By selecting only the most representative features, SFS outperformed
the baseline (90 features) by up to \SI{36}{\percent}, even though both
methods optimized the same number of features.

Crucially, SFS lowers computation without sacrificing accuracy.
Comparing the baseline (150 features) to SFS (90 features), the latter achieves
similar accuracy with a \SI{20}{\percent} reduction in optimization time
(\SI{53.7}{\milli\second} to \SI{43.0}{\milli\second}). The standard deviation
of the computation was also reduced by \SI{24.3}{\percent}. This demonstrates
that SFS can achieve similar accuracy with lower and more consistent
computational requirements.

\section{Demonstration on Legged Robots}
\begin{figure}
\centering
\vspace{0.5mm}
\includegraphics[width=0.85\columnwidth]{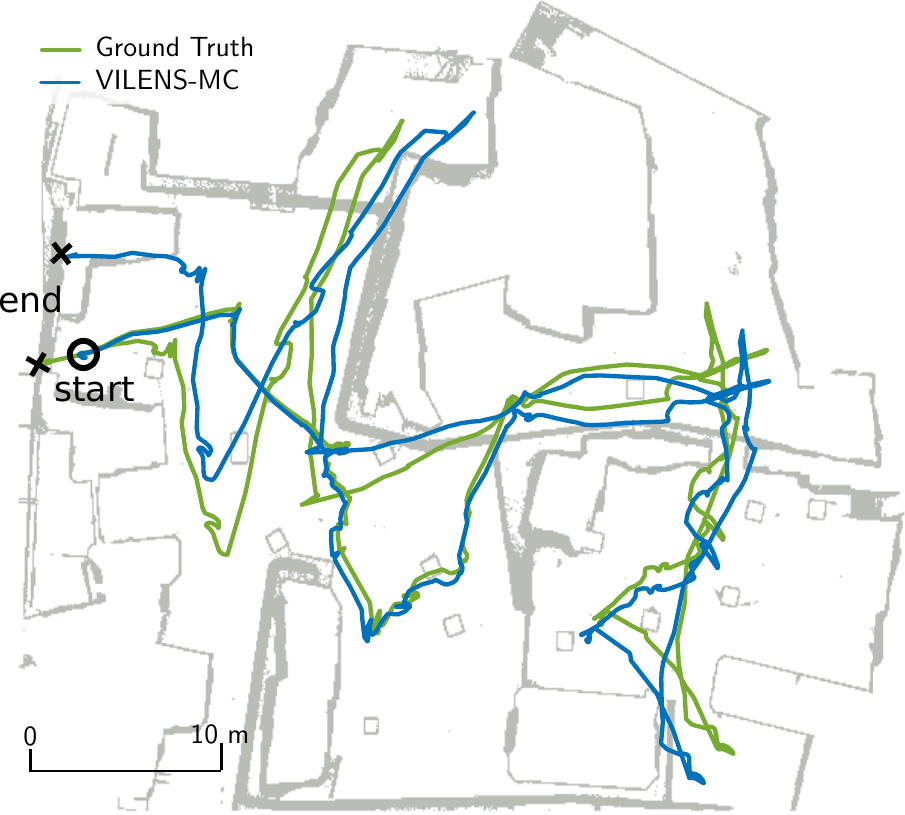}
\caption{State estimate on an ANYmal C quadruped robot autonomously exploring
an underground mine over $\sim$\SI{10}{\minute}. \Figure
\ref{fig:experimental-scenarios} shows the robot's on-board lighting which
caused strong illumination changes in the images.}

\label{fig:coyote-mine-trajectory}
\vspace{-3mm}
\end{figure}

To demonstrate the versatility of our approach, we tested VILENS-MC on the
ANYmal C quadruped robot equipped with an Alphasense for the
DARPA Subterranean Challenge \cite{Tranzatto2021}. In the
trajectory shown in \Figure \ref{fig:coyote-mine-trajectory} the robot
autonomously explored the same environment as the MINE dataset. VILENS-MC
achieved almost the same accuracy as for the handheld
dataset~(\SI{0.27}{\metre}, \SI{2.65}{\degree} RPE), indicating that
performance was similar across the different platforms.

\section{Conclusion}
\label{sec:conclusions}
We presented a novel factor graph formulation for state estimation that
tightly fuses an arbitrary number of cameras. The joint
optimization of IMU measurements with monocular and camera constraints
enables graceful handling of degenerate scenarios without requiring hard
switching between cameras. This also simplifies the initialization of the
individual monocular cameras. We have demonstrated comparable tracking
performance to state-of-the-art VIO systems in benign
conditions and better performance in extreme situations, such as in the case of
aggressive motions, loss of tracking in one or more cameras, or abrupt light
changes.

We proposed two algorithmic components specific to multi-camera odometry: cross
camera feature tracking and submatrix feature selection. Ablation studies
showed that these components improve accuracy and reduce optimization time.

Overall, we have presented a robust MC-VIO system that is capable of handling
various challenging real-world environments while consistently achieving
accurate state estimation. Future work would consider including camera
extrinsics into the optimization.



\bibliographystyle{./IEEEtran}
\bibliography{./IEEEabrv,./library}

\end{document}

%% file: shortcuts.tex
\newcommand{\hide}[1]{}
\newcommand{\Figure}{Fig.~}
\newcommand{\Equation}{Eq.~}

\newcommand{\ie}{{i.e.,~}}
\newcommand{\eg}{{e.g.,~}}

\newcommand{\bdmath}{\begin{dmath}}
\newcommand{\edmath}{\end{dmath}}
\newcommand{\beq}{\begin{equation}}
\newcommand{\eeq}{\end{equation}}
\newcommand{\bdm}{\begin{displaymath}}
\newcommand{\edm}{\end{displaymath}}
\newcommand{\bea}{\begin{eqnarray}}
\newcommand{\eea}{\end{eqnarray}}
\newcommand{\beal}{\beq \begin{array}{ll}}
\newcommand{\eeal}{\end{array} \eeq}
\newcommand{\beas}{\begin{eqnarray*}}
\newcommand{\eeas}{\end{eqnarray*}}
\newcommand{\ba}{\begin{array}}
\newcommand{\ea}{\end{array}}
\newcommand{\bit}{\begin{itemize}}
\newcommand{\eit}{\end{itemize}}
\newcommand{\ben}{\begin{enumerate}}
\newcommand{\een}{\end{enumerate}}

\newcommand{\SO}{\mathrm{SO}}

\newcommand{\Real}{\mathbb{R}}

\newcommand{\SEthree}{\ensuremath{\mathrm{SE}(3)}\xspace}

\newcommand{\SOthree}{\ensuremath{\SO(3)}\xspace}

\newcommand{\calC}{{\cal C}}

\newcommand{\calI}{{\cal I}}

\newcommand{\calN}{{\cal N}}

\newcommand{\calX}{{\cal X}}
\newcommand{\calZ}{{\cal Z}}

\newcommand{\T}{\mathbf{T}}
\newcommand{\R}{\mathbf{R}}

\newcommand{\Identity}{\mathbf{I}}

\newcommand{\transpose}{\mathsf{T}}

\newcommand{\tran}{\mathbf{p}}

\newcommand{\vel}{\mathbf{v}}

\newcommand{\bias}{\mathbf{b}}

\newcommand{\State}{\boldsymbol{x}}

\newcommand{\World}{\mathtt{W}}
\newcommand{\Imu}{\mathtt{I}}
\newcommand{\Camera}{\mathtt{C}}

\newcommand{\world}{\mathtt{{W}}}

\newcommand{\imu}{\mathtt{{I}}}
\newcommand{\base}{\mathtt{{B}}}

\newcommand{\Base}{\mathtt{{B}}}

\newcommand{\meters}{\rm{m}}

\newcommand{\pixel}{\mathbf{z}}

\newcommand{\Landmark}{\boldsymbol{m}}
\newcommand{\Landmarks}{\mathbf{M}}

\newcommand{\Measurement}{\boldsymbol{z}}
\newcommand{\Measurements}{\mathbf{Z}}

\newcommand{\hjac}{\mathbf{H}}

